\documentclass[submission,copyright,creativecommons]{eptcs}
\providecommand{\event}{ICLP 2023} 

\usepackage{ifpdf}

\ifpdf
  \usepackage{underscore}         
  \usepackage[T1]{fontenc}        
\else
  \usepackage{breakurl}           
\fi

\usepackage{listings}
\usepackage[english]{babel}
\usepackage[inline]{enumitem} 
\usepackage{csquotes}
\usepackage{amsmath}
\usepackage{amsthm}
\usepackage{tikz}
\usepackage{csvsimple-l3}
\usepackage{booktabs}
\usepackage[sort,compress]{cite}	
\usepackage[capitalize]{cleveref}

\newtheorem{definition}{Definition}

\newcommand{\mi}[1]{\ensuremath{\mathit{#1}}}

\newcommand{\slv}[1]{\textsc{#1}} 
\newcommand{\alphaslv}{\slv{Alpha}}

\newcommand{\heudirstmt}{\mathtt{\#heuristic}}

\newcommand{\heuhead}{\ensuremath{\mi{ha}}}
\newcommand{\heubody}{\ensuremath{\mi{hB}}}

\newcommand{\mytitle}[0]{Inductive Learning of\\Declarative Domain-Specific Heuristics for ASP}
\def\titlerunning{Inductive Learning of Declarative Domain-Specific Heuristics for ASP}
\def\authorrunning{R.\ Comploi-Taupe}

\title{\mytitle\thanks{
		This paper is a significant revision of a previous workshop paper \cite{DBLP:conf/lpnmr/Comploi-Taupe22}.
		Improvements include:
		Using \slv{FastLAS} instead of \slv{ilasp};
		creating the mode bias in a systematic way;
		using multiple examples in the learning task;
		learning \enquote{soft} in addition to \enquote{hard} heuristics;
		experimentally comparing learned heuristics to \slv{clingo}'s built-in heuristics as well as human-made heuristics;
		updated related work.
	}
}

\author
{
    Richard Comploi-Taupe\textsuperscript{1}
    \institute{
    \textsuperscript{1} Siemens AG Österreich, Vienna, Austria,
            \email{richard.taupe@siemens.com}
	}
}

\begin{document}
\label{firstpage}

\maketitle

\begin{abstract}
  Domain-specific heuristics are a crucial technique for the efficient solving of problems that are large or computationally hard.
  Answer Set Programming (ASP) systems support declarative specifications of domain-specific heuristics to improve solving performance.
  However, such heuristics must be invented manually so far.
  Inventing domain-specific heuristics for answer-set programs requires expertise with the domain under consideration and familiarity with ASP syntax, semantics, and solving technology.
  The process of inventing useful heuristics would highly profit from automatic support.
  This paper presents a novel approach to the automatic learning of such heuristics.
  We use Inductive Logic Programming (ILP) to learn declarative domain-specific heuristics from examples stemming from (near-)optimal answer sets of small but representative problem instances.
  Our experimental results indicate that the learned heuristics can improve solving performance and solution quality when solving larger, harder instances of the same problem.
\end{abstract}


\section{Introduction}
\label{sec:intro}

Answer Set Programming (ASP) \cite{aspbookgelfond,aspbooklifschitz,aspbookgebser,aspbookbaral}
is a declarative problem-solving approach applied successfully in many industrial and scientific domains.
For large and complex problems, however, domain-specific heuristics may be needed to achieve satisfactory performance \cite{DBLP:journals/tplp/DodaroGLMRS16,DBLP:journals/ki/FalknerFSTT18}.

Therefore, state-of-the-art ASP systems offer ways to integrate domain-specific heuristics in the solving process.
An extension for \slv{wasp} \cite{DBLP:conf/lpnmr/AlvianoADLMR19} facilitates external \emph{procedural} heuristics consulted at specific points during the solving process via an API \cite{DBLP:journals/tplp/DodaroGLMRS16}.
\emph{Declarative} specifications of domain-specific heuristics in the form of so-called \emph{heuristic directives} are supported by \slv{clingo} \cite{DBLP:journals/tplp/GebserKKS19,DBLP:conf/aaai/GebserKROSW13,potasscoguide} and \alphaslv\ \cite{DBLP:conf/lpnmr/Weinzierl17,DBLP:journals/jair/Comploi-TaupeFS23}.

However, such heuristics must be invented manually so far.
Human domain experts and ASP experts are needed to invent suitable domain-specific heuristics.
This paper presents a novel approach capable of learning basic declarative heuristics automatically.

Our core idea is to use Inductive Logic Programming (ILP) to learn declarative domain-specific heuristics from examples stemming from (near-)optimal answer sets of small but representative problem instances.
These heuristics can then be used to improve solving performance and solution quality for larger, harder problem instances.
Our experimental results are promising, indicating that this goal can be achieved.

After covering preliminaries in \cref{sec:prelim}, we present our main contribution in \cref{sec:maincontrib}.
\Cref{sec:experiments} presents experimental results, and \cref{sec:related} describes related work.
\Cref{sec:conclusions} concludes the paper by giving an outlook on future work.

\section{Preliminaries}
\label{sec:prelim}

In this section, we introduce a running example and cover preliminaries on domain-specific heuristics in ASP and inductive learning in ASP.
We assume familiarity with ASP and refer to \cite{aspbookgelfond,aspbooklifschitz,aspbookgebser,aspbookbaral} for detailed introductions.

\subsection{Running Example: The House Reconfiguration Problem (HRP)}
\label{sec:prelim:hrp}

The House Reconfiguration Problem (HRP) \cite{DBLP:conf/confws/FriedrichRFHSS11} is an abstracted version of industrial (re)configuration problems, e.g., rack configuration.
A complete description is available from the ASP Challenge 2019,\footnote{\url{https://sites.google.com/view/aspcomp2019/problem-domains}} and an encoding is available in Anna Ryabokon's PhD thesis
\cite{Ryabokon.2015}.

Formally, HRP is defined as a modification of the House Configuration Problem (HCP) \cite{DBLP:journals/jair/Comploi-TaupeFS23}.
\begin{definition}[HCP]
	The input for the \emph{House Configuration Problem (HCP)} is given by four sets of constants $P$, $T$, $C$, and $R$ representing persons, things, cabinets, and rooms, respectively, and an ownership relation $\mi{PT} \subseteq P \times T$ between persons and things. 
	
	The task is to find an assignment of things to cabinets $\mi{TC} \subseteq T \times C$ and cabinets to rooms $\mi{CR} \subseteq C \times R$ such that:
	\begin{enumerate*}[label=(\arabic*)]
		\item each thing is stored in a cabinet; 
		\item a cabinet contains at most five things; 
		\item every cabinet is placed in a room; 
		\item a room contains at most four cabinets; and 
		\item a room may only contain cabinets storing things of one person.
	\end{enumerate*}
\end{definition}
\begin{definition}[HRP]
	The input for the \emph{House Reconfiguration Problem (HRP)} is given by an HCP instance $H = \langle P, T, C, R, PT \rangle$, a legacy configuration $\langle \mi{TC}', \mi{CR}' \rangle$, and a set of things $T' \subseteq T$ that are defined as \enquote{long} (all other things are \enquote{short}).
	
	The task is then to find an assignment of things to cabinets $\mi{TC} \subseteq T \times C$ and cabinets to rooms $\mi{CR} \subseteq C \times R$, that satisfies all requirements of HCP as well as the following ones:
	\begin{enumerate*}[label=(\arabic*)]
		\item a cabinet is either small or high;
		\item a long thing can only be put into a high cabinet;
		\item a small cabinet occupies 1 and a high cabinet 2 of 4 slots
		available in a room;
		\item all legacy cabinets are small.
	\end{enumerate*}
\end{definition}

\begin{figure}
	\centering
\tikzstyle{person} = [draw, circle, inner sep=0em, minimum width=1.5em]
\tikzstyle{thing} =   [draw, rectangle, fill=white, inner sep=.2em, minimum width=2em, rounded corners=0.2em]
\tikzstyle{cabinet} =   [draw, rectangle, fill=black!20, inner sep=.2em, minimum width=2em]
\tikzstyle{room} =   [draw, rectangle, dashed, inner sep=.2em, minimum width=2em]

\begin{minipage}{.5\textwidth}
\begin{tikzpicture}[x=2em,y=2em,baseline=0pt]

\node[person] (p1) at (2.4, 1) {$\mathrm{p_1}$};
\node[person] (p2) at (6.0, 1) {$\mathrm{p_2}$};

\node[thing] (t1) at (0.0, -1) {$\mathrm{t_{1}}$};
\node[thing] (t2) at (1.2, -1) {$\mathrm{t_{2}}$};
\node[thing] (t3) at (2.4, -1) {$\mathrm{t_{3}}$};
\node[thing] (t4) at (3.6, -1) {$\mathrm{t_{4}}$};
\node[thing] (t5) at (4.8, -1) {$\mathrm{t_{5}}$};
\node[thing] (t6) at (6.0, -1) {$\mathrm{t_{6}}$};

\node[cabinet] (c1) at (8.0, 1) {$\mathrm{c_{1}}$};
\node[cabinet] (c2) at (8.0, 0.4) {$\mathrm{c_{2}}$};

\node[room] (r1) at (8.0, -0.4) {$\mathrm{r_{1}}$};
\node[room] (r2) at (8.0, -1) {$\mathrm{r_{2}}$};

\draw 
(p1)   -- (t1)
(p1)   -- (t2)
(p1)   -- (t3)
(p1)   -- (t4)
(p1)   -- (t5)
(p2)   -- (t6);
\end{tikzpicture}%
\end{minipage}%
\hfill\vline\hfill
\begin{minipage}{.4\textwidth}
\begin{tikzpicture}[x=2em,y=2em,baseline=0pt]

\draw[room] (-2, -0.6) rectangle node[left, label=left:$\mathrm{r_{1}}~~$]{} ++(2.6,3.5);
\draw[cabinet] (-1.2, -0.5) rectangle node[below, label=left:$\mathrm{c_{1}}$]{} ++(1.4,3.3);

\draw[room] (2, -0.6) rectangle node[left, label=left:$\mathrm{r_{2}}~~$]{} ++(2.6,3.5);
\draw[cabinet] (2.8, -0.5) rectangle node[below, label=left:$\mathrm{c_{2}}$]{} ++(1.4,3.3);

\node[thing] (t1) at (-0.5, 0.0) {$\mathrm{t_{1}}$};
\node[thing] (t2) at (-0.5, 0.6) {$\mathrm{t_{2}}$};
\node[thing] (t3) at (-0.5, 1.2) {$\mathrm{t_{3}}$};
\node[thing] (t4) at (-0.5, 1.8) {$\mathrm{t_{4}}$};
\node[thing] (t5) at (-0.5, 2.4) {$\mathrm{t_{5}}$};

\node[thing] (t6) at (3.5, 1.2) {$\mathrm{t_{6}}$};

\end{tikzpicture}
\end{minipage}
	\caption{Sample HRP instance (left) and one of its solutions (right) \protect\cite{DBLP:journals/jair/Comploi-TaupeFS23}}
	\label{fig:hrp-example}
\end{figure}
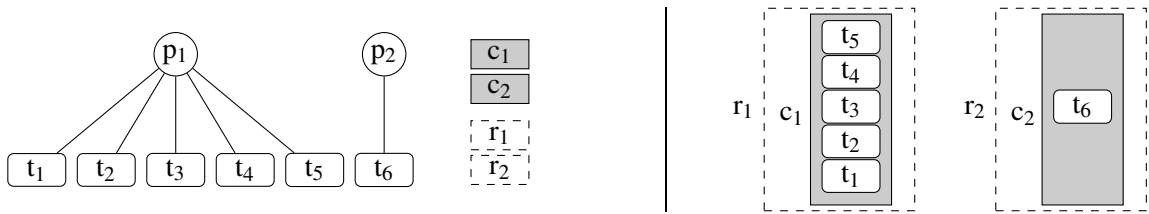

The sample HRP instance shown in \cref{fig:hrp-example} comprises two cabinets, two rooms, five things belonging to person $\mathrm{p_1}$, and one thing belonging to person $\mathrm{p_2}$. 
A legacy configuration is empty, and all things are small. 
In a solution, the first person's things are placed in cabinet $\mathrm{c_1}$ in the first room, and the thing of the second person is in cabinet $\mathrm{c_2}$ in the second room. 
For this sample instance, a solution of HRP corresponds to a solution of HCP \cite{DBLP:journals/jair/Comploi-TaupeFS23}.

Instances use the following predicates:
\texttt{cabinetDomainNew/1} and \texttt{roomDomainNew/1} define potential cabinet and potential rooms;
\texttt{thingLong/1} defines which things are long;
and \texttt{legacyConfig/1} defines all the other data in the legacy configuration,
e.g., \texttt{legacyConfig(personTOthing(p1,t1))} defines that person \texttt{p1} owns thing \texttt{t1}, and \texttt{legacyConfig(roomTOcabinet(r1,c1))} specifies one tuple in the legacy assignment of cabinets to rooms.

For the present work, unique predicates have been used instead of predicates with function terms, e.g., \texttt{legacyConfig\_personTOthing(p1,t1)} instead of \texttt{legacyConfig(personTOthing(p1,t1))}.

The main two choice rules guessing the assignment of things to cabinets and the assignment of cabinets to rooms look as follows:
\begin{lstlisting}[numbers=none]
	{ cabinetTOthing(C,T) } :- cabinetDomain(C), thing(T).
	{ roomTOcabinet(R,C) }  :- roomDomain(R),  cabinet(C).
\end{lstlisting}

To define the domains of cabinets and rooms as the union of existing objects and newly available identifiers, the encoding also contains the following rules:
\begin{lstlisting}[numbers=none]
	cabinetDomain(C) :- cabinetDomainNew(C).
	cabinetDomain(C) :- legacyConfig_cabinet(C).
	roomDomain(R)    :- roomDomainNew(R).
	roomDomain(R)    :- legacyConfig_room(R).
\end{lstlisting}

Instances may optionally include atoms of various predicates to define \emph{costs} for specific actions such as placing a thing in a cabinet, placing a cabinet in a room, reusing an existing placement of a thing in a cabinet, reusing an existing placement of a cabinet in a room, removing a thing from a cabinet, removing a cabinet from a room, etc.
These costs are then determined based on the difference between solution $\langle \mi{TC}, \mi{CR} \rangle$ and legacy configuration $\langle \mi{TC}', \mi{CR}' \rangle$.
A weak constraint in the encoding instructs the solver to minimise the costs.

Each available instance belongs to one of four instance classes \cite{DBLP:conf/confws/FriedrichRFHSS11,Ryabokon.2015}:
\textit{Empty} (\enquote{ec}, empty legacy configuration); \textit{long} (\enquote{lt}, some things are long); \textit{new room} (\enquote{nr}, some cabinets have to be reallocated to new rooms); and \textit{swap} (\enquote{ss}, only one person and a specific pattern of legacy configuration).

\subsection{Domain-Specific Heuristics in ASP}
\label{sec:prelim:domheu}

To solve large instances of industrial problems, employing an ASP solver out-of-the-box may not be sufficient.
Sophisticated encodings or solver tuning methods (such as portfolio solving) are common ways to deal with this issue.

Domain-specific heuristics are another way to speed-up answer set solving.
They were even needed to achieve breakthroughs in solving industrial configuration problems with ASP \cite{DBLP:journals/tplp/DodaroGLMRS16}.

Several approaches have been proposed to embed heuristic knowledge into the ASP solving process.
\slv{hwasp} \cite{DBLP:journals/tplp/DodaroGLMRS16} extends \slv{wasp} \cite{DBLP:conf/lpnmr/AlvianoADLMR19} by facilitating external procedural heuristics consulted at specific points during the solving process via an API.

A declarative approach to formulating domain-specific heuristics in ASP is provided by \slv{clingo},\footnote{\url{https://potassco.org/clingo/}} supporting $\heudirstmt$ directives \cite{DBLP:conf/aaai/GebserKROSW13,potasscoguide}.
Heuristic directives enable the declarative specification of weights determining atom and sign orders in a solver's internal decision heuristics.
An atom's weight influences the order in which atoms are considered by the solver when making a decision.
A sign modifier instructs whether the selected atom must be assigned true or false.
Atoms with a higher weight are assigned a value before atoms with a lower weight.

In the syntax for (non-ground) heuristic directives in \slv{clingo} \labelcref{eq:heudir}, $\heuhead$ is an atom, $\heubody$ is a conjunction of literals representing the heuristic body, and $w$, $p$, and $m$ are terms \cite{potasscoguide}.
\begin{align}
	\label{eq:heudir}
	&\#heuristic ~~ \heuhead : \heubody. \qquad [w@p,m]&
\end{align}
The optional term $p$ gives a preference between heuristic values for the same atom (preferring those with higher $p$).
The term $m$ specifies the type of heuristic information and can take the following values: \texttt{sign}, \texttt{level}, \texttt{true}, \texttt{false}, \texttt{init} and \texttt{factor}.
For instance, heuristics for $m{=}$\texttt{init} and $m{=}$\texttt{factor} allow modifying initial and actual atom scores evaluated by the solver's decision heuristics (e.g., VSIDS).
The $m{=}$\texttt{sign} modifier forces the decision heuristics to assign an atom $\heuhead$ a specific sign, i.e., true or false, and $m{=}$\texttt{level} allows for the definition of an order in which the atoms are assigned---the larger the value of $w$, the earlier an atom must be assigned. 
Finally, $m{=}$\texttt{true} specifies that $a$ should be assigned true with weight $w$ if $\heubody$ is satisfied, and $m{=}$\texttt{false} is the analogue heuristics that assigns $a$ false.

A new approach implemented in the lazy-grounding ASP system \slv{Alpha},\footnote{\url{https://github.com/alpha-asp/Alpha}} based on the \slv{clingo} approach, has introduced novel semantics for heuristic directives aimed at non-monotonic heuristics \cite{DBLP:journals/jair/Comploi-TaupeFS23}.

\subsection{Inductive Learning in ASP}
\label{sec:prelim:ilasp}

Inductive Logic Programming (ILP) is an approach to learning a program that explains a set of examples given some background knowledge.
\slv{FastLAS}\footnote{\url{https://spike-imperial.github.io/FastLAS/}} \cite{DBLP:conf/aaai/LawRBB020,DBLP:conf/ijcai/LawRBB21} is a system capable of learning Answer Set Programs.\footnote{\slv{FastLAS} is not the only ILP system in the ASP setting. It was chosen for this work because of its speed and recency. \slv{ilasp} \cite{DBLP:journals/corr/abs-2005-00904}, on the other hand, has successfully been used in a previous version of the paper \cite{DBLP:conf/lpnmr/Comploi-Taupe22}.}

\slv{FastLAS} operates on a \emph{learning task}, which consists of three components \cite{DBLP:conf/aaai/LawRBB020,DBLP:conf/ijcai/LawRBB21}:
The \emph{background knowledge} $B$ (an ASP program already known before learning), the \emph{mode bias} $M$ (that expresses which ASP programs can be learned), and the \emph{examples} $E$ (which specify properties the learned program must satisfy).
When the properties specified by a particular example in $E$ are satisfied, the example is said to be \emph{covered}.

\slv{FastLAS} finds a program (often called a hypothesis) $H$ such that $B \cup H$ covers every example in $E$ (or, if the examples are considered \emph{noisy}, such that the total penalty of non-covered examples is minimised) \cite{DBLP:conf/ijcai/LawRBB21}.
$H$ is an element of the search space defined by $M$.

The syntax used to define learning tasks for \slv{FastLAS} follows the one used by \slv{ilasp} (cf.\ \cite{DBLP:journals/corr/abs-2005-00904,ilasp_manual}).

\subsubsection{Mode Bias}
\label{sec:prelim:m}

The mode bias consists of a set of \emph{mode declarations}
of types
\texttt{\#modeh} and \texttt{\#modeb}, specifying what the heads and bodies of learned rules may look like, respectively.
A \emph{placeholder} is a term \texttt{var(t)} or \texttt{const(t)} for some constant term \texttt{t}.
Such placeholders can be replaced by any variable or constant (respectively) of \emph{type} \texttt{t} \cite{DBLP:conf/ijcai/LawRBB21}.

As a simple example, \cref{lst:m} shows part of the mode bias for a learning task for the HRP.
\lstinputlisting[label=lst:m,caption={Part of the mode bias for HRP},linerange={1-3}]{listings/example\_for\_paper.las}
The first mode declaration specifies that the binary predicate \texttt{cabinetTOthing} can be used in the head of rules and that its terms are of variable types \texttt{cabinetDomain} and \texttt{thing} (in this order).
The other two mode declarations specify which predicates can occur in the bodies of learned rules.
Note that the same terms may be used in learned rules wherever the same placeholders are used.

\pagebreak\noindent
Thus, the rule space defined by the mode bias given in \cref{lst:m} consists of the following rule:
\begin{lstlisting}[numbers=none]
	cabinetTOthing(V0,V1) :- cabinetDomain(V0), thing(V1).
\end{lstlisting}
\slv{FastLAS} (contrary to \slv{ilasp} \cite{DBLP:journals/corr/abs-2005-00904,law_2022}) uses strict types, which need to be defined \cite{DBLP:conf/aaai/LawRBB020}.
Thus, the rules in \cref{lst:b} are needed in the background knowledge to make the mode bias from \cref{lst:m} work.
\lstinputlisting[label=lst:b,caption={Definition of some strict types for HRP},linerange={12-16}]{listings/example\_for\_paper.las}

\subsubsection{Examples}
\label{sec:prelim:e}

A positive example is given by a \texttt{\#pos} statement, and a negative example by a \texttt{\#neg} statement \cite{DBLP:journals/corr/abs-2005-00904,ilasp_manual}.
Each example consists of several components:\footnote{Currently, we do not use penalties.}
an optional example identifier, a set of ground atoms called \emph{inclusions}, a set of ground atoms called \emph{exclusions}, and an optional set of rules (usually just facts) called \emph{context}.

A positive example is covered iff there exists at least one answer set for $B \cup H$ that contains all of the inclusions and none of the exclusions.
A negative example
states that there must \emph{not} exist an answer set that contains all of its inclusions and none of its exclusions \cite{DBLP:journals/corr/abs-2005-00904,ilasp_manual}.

The \emph{context} is the problem instance to which the inclusions and exclusions refer (considering the usual distinction between unvarying problem encoding and problem instances specified by facts) \cite{DBLP:journals/corr/abs-2005-00904,ilasp_manual}.

\Cref{lst:e} shows a simplified example for HRP stating that \texttt{cabinetTOthing(1,2)} shall be true for the problem instance in which \texttt{cabinetDomainNew(1)} and \texttt{thing(2)} are true.
The identifier of this example is \texttt{ex1}.
\lstinputlisting[label=lst:e,caption={A simplified example in a learning task for HRP},linerange={5-9}]{listings/example\_for\_paper.las}

\section{Inductive Learning of Domain-Specific Heuristics}
\label{sec:maincontrib}

This section presents the main contribution of this paper---our approach to the inductive learning of domain-specific heuristics for ASP.

The basic idea is to solve one or more small but representative instances of a problem, use the resulting answer sets as positive examples for inductive learning, learn a set of definite rules, and transform the learned rules into declarative heuristic directives in the form of \cref{eq:heudir} presented in \cref{sec:prelim:domheu}.
These heuristics can then be used to speed up solving larger/harder instances of the same problem.

The learning task for \slv{FastLAS} is defined in our approach in the way described in the following paragraphs.

\paragraph{Head mode declarations.}
Predicates that appear in the problem encoding in the heads of choice elements in the heads of choice rules\footnote{More formally: The head of a choice rule contains a collection of choice elements, each of the form $a \colon l_1, \dots, l_k$ \cite{DBLP:journals/tplp/CalimeriFGIKKLM20}; the predicates used for $a$ are the ones that are used to create \texttt{\#modeh} declarations.} are used to create \texttt{\#modeh} declarations.
The placeholders are created in the following way:
\begin{itemize}
	\item If the rule body and the choice condition together are (indirectly via rules) completely determined by the problem instance, use the body/condition predicate name as a strict type name and include the defining rules in the background knowledge.
	For example, from the choice rule in line 1 in \cref{lst:hrp:enc}, the mode bias in line 2 in \cref{lst:hrp:las} along with the background knowledge in lines 11--12 in \cref{lst:hrp:las} is created.
	\item If the head predicate depends on another predicate that is used as a \texttt{\#modeh} predicate itself, use the underlying strict type.
	Example: Line 2 in \cref{lst:hrp:enc}, lines 3--4 in \cref{lst:hrp:las}.
	\item To define placeholders for non-unary predicates, define strict types for each argument in a deterministic way, appending \texttt{\_arg1} etc.\ to the predicate name, along with corresponding rules.
	Example: Line 3 in \cref{lst:hrp:enc}, lines 5--6 and 13--14 in \cref{lst:hrp:las}.
\end{itemize}

\paragraph{Body mode declarations.}
All other predicates appearing in the problem encoding are used to create \texttt{\#modeb} declarations.
Placeholders are directly built from the predicate names for unary predicates, and with suffixes such as \texttt{\_arg1} etc.\ as above for non-unary predicates.
As an example, cf.\ \cref{lst:hrp:enc} and lines 8--9 in \cref{lst:hrp:las}.

\paragraph{Background knowledge.}
The background knowledge consists only of rules needed to define strict types as described above.

\paragraph{Examples.}
As positive examples for learning, answer sets for small but representative problem instances are used (one answer set per problem instance, yielding one example per problem instance).
In case the underlying problem is an optimisation problem (like the HRP described in \cref{sec:prelim:hrp}), we propose to use (near-)optimal answer sets for this process.
The (yet unproven) hypothesis is that learning from better answer sets yields better heuristics.

We use context-dependent examples; the context is given by the problem instance.
The set of inclusions corresponds to the answer set filtered to cover only the predicates appearing in \texttt{\#modeh}, and the set of exclusions is empty.

\subsection{Learning in the House Reconfiguration Problem}
\label{sec:maincontrib:example}

Let us re-consider the running example from \cref{sec:prelim:hrp}, the House Reconfiguration Problem (HRP).
As representative problem instances, the smallest instance of each of the four instance classes was used.
A near-optimal\footnote{\slv{clingo} was used to find the best solution that could be found with human-made domain-specific heuristics within 10 minutes; for some instances, optimality could not be proven within several days of search.} answer set for each of these instances was computed by \slv{clingo} \cite{DBLP:journals/tplp/GebserKKS19}.

We built a learning task from the full HRP encoding and the four representative instances as described at the beginning of \cref{sec:maincontrib}.
The entire mode bias contains 16 \texttt{\#modeh} declarations and 32 \texttt{\#modeb} declarations and is too large to be shown here.
\Cref{lst:hrp:las} shows the parts of the learning task that are created for the choice rules in \cref{lst:hrp:enc}.
\pagebreak
\lstinputlisting[label=lst:hrp:enc,caption={Some choice rules from the HRP encoding},linerange={10-10,52-52,69-69}]{listings/normalised\_house\_encoding\_aspcore2.asp}
\lstinputlisting[label=lst:hrp:las,caption={Part of the learning task for HRP},linerange={1-2,6-9,19-19,21-21,31-31,3026-3028,3035-3036}]{listings/house\_FastLAS\_v4\_for\_paper.las}

\paragraph{Results.}
The following rules form the hypothesis learned by \slv{FastLAS}:\footnote{\slv{FastLAS} was called with parameter \texttt{-{}-force-safety} to enforce learned rules to be save also without artificially added strict type atoms, and those atoms have been stripped from the learned rules in postprocessing where they were redundant. With this learning task, it does not seem to make a difference whether \slv{FastLAS} is called with \texttt{-{}-opl} or \texttt{-{}-nopl}. Rules have been re-ordered manually to improve readability.}
\begin{lstlisting}[numbers=none]
cabinet(V0)       :- cabinetDomain(V0).
cabinetHigh(V0)   :- cabinetDomain(V0).
cabinetSmall(V0)  :- cabinetDomain(V0).
room(V0)          :- roomDomain(V0).
roomTOcabinet(V0,V1)  :- roomDomain(V0), cabinetDomain(V1).
cabinetTOthing(V0,V1) :- cabinetDomain(V0), thing(V1).
reuse_room(V0)        :- legacyConfig_room(V0).
reuse_cabinet(V0)     :- legacyConfig_cabinet(V0).
reuse_personTOroom(V0,V1)    :- legacyConfig_personTOroom(V0,V1).
reuse_roomTOcabinet(V0,V1)   :- legacyConfig_roomTOcabinet(V0,V1).
delete_roomTOcabinet(V0,V1)  :- legacyConfig_roomTOcabinet(V0,V1).
reuse_cabinetTOthing(V0,V1)  :- legacyConfig_cabinetTOthing(V0,V1).
delete_cabinetTOthing(V0,V1) :- legacyConfig_cabinetTOthing(V0,V1).
\end{lstlisting}
\slv{FastLAS} needed approximately 3 minutes to come up with this result on the author's personal computer.

As a next step, we transformed these rules to heuristics in the form of \cref{eq:heudir} by using the head of each rule as $\heuhead$ and the body as $\heubody$.
Syntactically, this replaces \texttt{:-} with \texttt{:} in each rule and adds an appropriate annotation at the end of each resulting heuristic directive.
The learned heuristics instruct the solver to choose the heads of most--but not all--choice rules.

For the annotation, we tried two approaches.
The first annotation used for all heuristic directives is \texttt{[1,true]}.
The idea is to instruct the solver to make the heads of the heuristics true, since the positive examples consist of atoms that are true in answer sets.
And since we don't have any information on prioritising the heuristics at this point, all get the same weight 1.

The second approach uses the annotation \texttt{[2,factor]}.
By this, the solver will multiply the atom scores of its domain-independent heuristics by the factor 2 for the atoms indicated by the heuristic directives.
The idea here is to gently steer the solver into the right direction without enforcing any decisions.

\section{Experimental Results}
\label{sec:experiments}

To test the effects of the learned heuristics, we used \slv{clingo} version 5.6.2 to solve all available HRP instances (94 in number) with and without the learned heuristics.
Additionally, other (built-in and human-made) heuristics were used in the experiments for comparison.
The HRP instances stem from previous experiments \cite{DBLP:journals/jair/Comploi-TaupeFS23}.
These instances were generated in the pattern of the original instances \cite{DBLP:conf/confws/FriedrichRFHSS11}.
This pattern represents four different reconfiguration scenarios encountered in practice, and the instances are abstracted real-world instances.
Our instances are considerably larger than the original ones, though (ranging up to 800 things, while the original instances used at most 280 things).

Each of the machines used to run the experiments
ran Ubuntu 22.04.2 LTS Linux and
was equipped with two
Intel\textsuperscript{\textregistered} Xeon\textsuperscript{\textregistered} E5-2650 v4 @ 2.20GHz CPUs with 12 cores.
Hyperthreading was disabled and the maximum CPU frequency was set to 2.90GHz.
Scheduling of benchmarks was done with Slurm\footnote{\url{https://slurm.schedmd.com/}} version 21.08.5.
Runsolver\footnote{\url{https://github.com/utpalbora/runsolver}} v3.4.1 was used to limit time consumption to 10 minutes per instance and memory to 20 GiB.
Care was taken to avoid side effects between CPUs, e.g., by requesting exclusive access to an entire machine for each benchmark from Slurm.

\slv{clingo} was instructed to search for the optimal answer set in its default configuration, given an encoding including a weak constraint.
After 10 minutes per instance, search was aborted and the optimisation value of the best solution found so far was recorded.
The solver was used in the following five configurations:
\begin{description}
	\item[plain:] the plain encoding without any domain-specific heuristics
	\item[learned (hard):] the learned heuristics with annotation \texttt{[1,true]}, i.e., first assigning true on all atoms determined by the learned heuristics
	\item[learned (soft):] the learned heuristics with annotation \texttt{[2,factor]}, i.e., modifying the atom scores of the solver-internal heuristics by the factor 2 for all atoms determined by the learned heuristics
	\item[built-in:] \slv{clingo}'s built-in heuristics \texttt{-{}-dom-mod=false,opt}, i.e., preferring atoms being optimised
	\item[human-made:] the human-made heuristics introduced by Comploi-Taupe et al.\ \cite{DBLP:journals/jair/Comploi-TaupeFS23}.
\end{description}

\begin{table}
	\centering
	\caption{Experimental results: Achieved optimisation values without and with various heuristics, and relative improvement (positive percentage, boldface) or deterioration (negative percentage, italics)}
	\label{tab:results}
	\begin{filecontents*}{tables/results-clingo-prepared.csv}
instancefull,instance,clingoplain,clingoheuhard,clingoheusoft,clingoheubuiltin,clingoheuhuman,imprclingoheuhard,imprclingoheusoft,imprclingoheubuiltin,imprclingoheuhuman,outclingoplain,outclingoheuhard,outclingoheusoft,outclingoheubuiltin,outclingoheuhuman
ec-0001-house-t0100,ec-0001,100,125,100,$\infty$,100,-25\%,0\%,$-\infty$,0\%,100,\textit{125 (-25\%)},100 (0\%),\textit{$\infty$ ($-\infty$)},100 (0\%)
ec-0002-house-t0125,ec-0002,125,195,125,$\infty$,125,-56\%,0\%,$-\infty$,0\%,125,\textit{195 (-56\%)},125 (0\%),\textit{$\infty$ ($-\infty$)},125 (0\%)
ec-0003-house-t0150,ec-0003,155,230,150,$\infty$,150,-48\%,3\%,$-\infty$,3\%,155,\textit{230 (-48\%)},\textbf{150 (3\%)},\textit{$\infty$ ($-\infty$)},\textbf{150 (3\%)}
ec-0004-house-t0175,ec-0004,180,290,175,$\infty$,175,-61\%,3\%,$-\infty$,3\%,180,\textit{290 (-61\%)},\textbf{175 (3\%)},\textit{$\infty$ ($-\infty$)},\textbf{175 (3\%)}
ec-0005-house-t0200,ec-0005,230,335,200,$\infty$,200,-46\%,13\%,$-\infty$,13\%,230,\textit{335 (-46\%)},\textbf{200 (13\%)},\textit{$\infty$ ($-\infty$)},\textbf{200 (13\%)}
ec-0006-house-t0225,ec-0006,270,675,225,$\infty$,225,-150\%,17\%,$-\infty$,17\%,270,\textit{675 (-150\%)},\textbf{225 (17\%)},\textit{$\infty$ ($-\infty$)},\textbf{225 (17\%)}
ec-0007-house-t0250,ec-0007,310,550,250,$\infty$,250,-77\%,19\%,$-\infty$,19\%,310,\textit{550 (-77\%)},\textbf{250 (19\%)},\textit{$\infty$ ($-\infty$)},\textbf{250 (19\%)}
ec-0008-house-t0275,ec-0008,350,1220,275,$\infty$,275,-249\%,21\%,$-\infty$,21\%,350,\textit{1220 (-249\%)},\textbf{275 (21\%)},\textit{$\infty$ ($-\infty$)},\textbf{275 (21\%)}
ec-0009-house-t0300,ec-0009,385,1250,505,$\infty$,300,-225\%,-31\%,$-\infty$,22\%,385,\textit{1250 (-225\%)},\textit{505 (-31\%)},\textit{$\infty$ ($-\infty$)},\textbf{300 (22\%)}
ec-0010-house-t0325,ec-0010,1045,1585,1045,$\infty$,325,-52\%,0\%,$-\infty$,69\%,1045,\textit{1585 (-52\%)},1045 (0\%),\textit{$\infty$ ($-\infty$)},\textbf{325 (69\%)}
ec-0011-house-t0350,ec-0011,1340,1750,1325,$\infty$,350,-31\%,1\%,$-\infty$,74\%,1340,\textit{1750 (-31\%)},\textbf{1325 (1\%)},\textit{$\infty$ ($-\infty$)},\textbf{350 (74\%)}
lt-0001-house-t0180,lt-0001,337,980,460,$\infty$,192,-191\%,-36\%,$-\infty$,43\%,337,\textit{980 (-191\%)},\textit{460 (-36\%)},\textit{$\infty$ ($-\infty$)},\textbf{192 (43\%)}
lt-0002-house-t0210,lt-0002,1347,1138,1330,$\infty$,224,16\%,1\%,$-\infty$,83\%,1347,\textbf{1138 (16\%)},\textbf{1330 (1\%)},\textit{$\infty$ ($-\infty$)},\textbf{224 (83\%)}
lt-0003-house-t0240,lt-0003,1589,1302,1527,$\infty$,256,18\%,4\%,$-\infty$,84\%,1589,\textbf{1302 (18\%)},\textbf{1527 (4\%)},\textit{$\infty$ ($-\infty$)},\textbf{256 (84\%)}
lt-0004-house-t0270,lt-0004,1848,$\infty$,1798,$\infty$,288,$-\infty$,3\%,$-\infty$,84\%,1848,\textit{$\infty$ ($-\infty$)},\textbf{1798 (3\%)},\textit{$\infty$ ($-\infty$)},\textbf{288 (84\%)}
lt-0005-house-t0300,lt-0005,2103,$\infty$,1979,$\infty$,320,$-\infty$,6\%,$-\infty$,85\%,2103,\textit{$\infty$ ($-\infty$)},\textbf{1979 (6\%)},\textit{$\infty$ ($-\infty$)},\textbf{320 (85\%)}
lt-0006-house-t0330,lt-0006,2382,$\infty$,2357,$\infty$,352,$-\infty$,1\%,$-\infty$,85\%,2382,\textit{$\infty$ ($-\infty$)},\textbf{2357 (1\%)},\textit{$\infty$ ($-\infty$)},\textbf{352 (85\%)}
nr-0001-house-t0192,nr-0001,1297,1064,1283,$\infty$,256,18\%,1\%,$-\infty$,80\%,1297,\textbf{1064 (18\%)},\textbf{1283 (1\%)},\textit{$\infty$ ($-\infty$)},\textbf{256 (80\%)}
nr-0002-house-t0216,nr-0002,1124,1194,1403,$\infty$,288,-6\%,-25\%,$-\infty$,74\%,1124,\textit{1194 (-6\%)},\textit{1403 (-25\%)},\textit{$\infty$ ($-\infty$)},\textbf{288 (74\%)}
nr-0003-house-t0240,nr-0003,1711,1324,1550,$\infty$,320,23\%,9\%,$-\infty$,81\%,1711,\textbf{1324 (23\%)},\textbf{1550 (9\%)},\textit{$\infty$ ($-\infty$)},\textbf{320 (81\%)}
nr-0004-house-t0264,nr-0004,1023,$\infty$,1739,$\infty$,352,$-\infty$,-70\%,$-\infty$,66\%,1023,\textit{$\infty$ ($-\infty$)},\textit{1739 (-70\%)},\textit{$\infty$ ($-\infty$)},\textbf{352 (66\%)}
nr-0005-house-t0288,nr-0005,2008,$\infty$,1870,$\infty$,384,$-\infty$,7\%,$-\infty$,81\%,2008,\textit{$\infty$ ($-\infty$)},\textbf{1870 (7\%)},\textit{$\infty$ ($-\infty$)},\textbf{384 (81\%)}
nr-0006-house-t0312,nr-0006,$\infty$,$\infty$,2173,$\infty$,416,$\infty$,100\%,$\infty$,100\%,$\infty$,\textbf{$\infty$},\textbf{2173 (100\%)},\textbf{$\infty$},\textbf{416 (100\%)}
nr-0007-house-t0336,nr-0007,$\infty$,$\infty$,$\infty$,$\infty$,448,$\infty$,$\infty$,$\infty$,100\%,$\infty$,\textbf{$\infty$},\textbf{$\infty$},\textbf{$\infty$},\textbf{448 (100\%)}
ss-0001-house-t0175,ss-0001,46,176,4,4,4,-283\%,91\%,91\%,91\%,46,\textit{176 (-283\%)},\textbf{4 (91\%)},\textbf{4 (91\%)},\textbf{4 (91\%)}
ss-0002-house-t0210,ss-0002,209,207,8,4,4,1\%,96\%,98\%,98\%,209,\textbf{207 (1\%)},\textbf{8 (96\%)},\textbf{4 (98\%)},\textbf{4 (98\%)}
ss-0003-house-t0245,ss-0003,726,242,843,4,4,67\%,-16\%,99\%,99\%,726,\textbf{242 (67\%)},\textit{843 (-16\%)},\textbf{4 (99\%)},\textbf{4 (99\%)}
ss-0004-house-t0280,ss-0004,1580,707,1649,4,4,55\%,-4\%,100\%,100\%,1580,\textbf{707 (55\%)},\textit{1649 (-4\%)},\textbf{4 (100\%)},\textbf{4 (100\%)}
ss-0005-house-t0315,ss-0005,1785,840,1655,4,4,53\%,7\%,100\%,100\%,1785,\textbf{840 (53\%)},\textbf{1655 (7\%)},\textbf{4 (100\%)},\textbf{4 (100\%)}
ss-0006-house-t0350,ss-0006,2313,1272,2124,4,4,45\%,8\%,100\%,100\%,2313,\textbf{1272 (45\%)},\textbf{2124 (8\%)},\textbf{4 (100\%)},\textbf{4 (100\%)}
\end{filecontents*}
\csvreader[
	head to column names,
	range = 1-31,
	tabular = lrrrrr,
	table head = \toprule \bfseries Instance & \bfseries plain & \bfseries learned (hard) & \bfseries learned (soft) & \bfseries built-in & \bfseries human-made \\\midrule,
	late after line = \\\midrule,
	late after last line = \\,
	table foot = \bottomrule,
	]{tables/results-clingo-prepared.csv}{}{%
		\instance & \outclingoplain & \outclingoheuhard & \outclingoheusoft & \outclingoheubuiltin & \outclingoheuhuman
	}
\end{table}

\Cref{tab:results} shows the achieved optimisation values and the relative improvement when using the learned heuristics for all 30 instances that could be solved in any solver configuration.
For the other 64 instances, no answer set could be found in any configuration; therefore, they are not included in the table.

The first column shows the instance identifier.
The first two characters of each identifier refer to one of the four instance classes of the HRP (cf.~\cref{sec:prelim:hrp}).
The numeric part of the identifier increases with increasing instance size.

The second column contains the achieved optimisation values in the \enquote{plain} solver configuration, i.e., without domain-specific heuristics.
The remaining columns contain the achieved optimisation values using the other solver configurations (\enquote{learned (hard)}, \enquote{learned (soft)}, \enquote{built-in}, and \enquote{human-made}).
In all columns, the symbol $\infty$ is used instead of an optimisation value when no answer set could be found within the time limit of 10 minutes.

The columns for non-plain solver configurations additionally show, within parentheses, the respective change in the optimisation value when using heuristics relative to solving without domain-specific heuristics.
A positive percentage signifies an improvement (printed in boldface), and negative values indicate a deterioration (printed in italics).
The value 100\% is used in cases where an answer set could be found only when using one of the heuristics 
and the value $-\infty$ is used when an answer set could be found only without heuristics.

The learned heuristics seem to have positive effects even though they are (still) straightforward.
Positive effects could be seen especially when the heuristics were applied in a \enquote{soft} way, weighting but not replacing the domain-independent search heuristics,
leading to improved solution quality in 20 cases and to a deterioration in 6 cases.
However, improvements varied strongly between different instances.
Furthermore, results are sensitive to the chosen time-out.
For example, we observed stronger improvements (on fewer solved instances) when experimenting with a time-out of three minutes instead of ten.

Surprisingly, the solver benefited only rarely from learned \enquote{hard} heuristics as well as the built-in heuristics preferring atoms being optimised.
Human-made heuristics still outperformed learned heuristics by a great degree, which could show untapped potential for our approach.

Besides our experiments with \slv{clingo}, we also experimented with the lazy-grounding ASP system \alphaslv\ \cite{DBLP:conf/lpnmr/Weinzierl17}.
This system accepts heuristic directives in a slightly different syntax \cite{DBLP:journals/jair/Comploi-TaupeFS23}.
Without domain-specific heuristics, \alphaslv\ could solve none of the HRP instances under consideration.
The heuristics learned by our approach enabled \alphaslv\ to solve three instances (without optimisation, which is not yet supported by \alphaslv).
Human-made heuristics (cf.\ \cite{DBLP:journals/jair/Comploi-TaupeFS23}) enabled \alphaslv\ to solve 58 of these instances on the same benchmarking infrastructure.

\section{Related Work}
\label{sec:related}

Balduccini \cite{DBLP:journals/aicom/Balduccini11} has also presented an approach to learning domain-specific heuristics offline from representative instances.
The basic idea, which is very different to our approach, is to record which choices are made in the path of a search tree that led to a solution and to use this information to compute probabilities for decisions on ground atoms.
These probabilities are then used while solving other problem instances to reduce the likelihood of backtracks.
The approach is restricted to DPLL-style solvers like \slv{smodels} \cite{DBLP:conf/lpnmr/SyrjanenN01}, and extending it to CDCL-based systems like \slv{clingo} is mentioned as future work.

A similar approach is aimed at configuration problems encoded as constraint satisfaction problems (CSPs) \cite{jannach2013toward}.

Dodaro et al.\ \cite{DBLP:conf/lpnmr/DodaroIOR22} use deep learning to learn domain-specific heuristics for the graph colouring problem to be used by the solver \slv{wasp}.
As usual with deep learning, this approach requires huge numbers of training instances (in reported experiments, $210,000$ instances were used for training, 60\% of which have been randomly chosen to build the training set).
The approach, including the integration in \slv{wasp}'s solving algorithm, is specific to the graph colouring problem.
In contrast to the work presented in this paper, learned heuristics apply to ground ASP and don't offer a concise, declarative representation readable by humans.

\section{Conclusions and Future Work}
\label{sec:conclusions}

We have proposed a novel approach to inductively learning declarative specifications of domain-specific heuristics for ASP from answer sets of small but representative instances.
Our approach employs the inductive learning system \slv{FastLAS}.
Utilising an example representing a significant real-world configuration problem, we have demonstrated that simple heuristics can easily be learned.

Experimental results are promising, since the learned heuristics led to improved solution quality in many cases.
The fact that so far, we have only learned very simple heuristics and those already led to significant improvements is encouraging.
Future work will show whether our method can be extended to learn more complex heuristics that can improve solving performance and solution quality even further.

{
	\footnotesize
	\paragraph{Acknowledgements.}
	The author is grateful to the participants of the two workshops HYDRA 2022\footnote{\url{https://hydra2022.demacs.unical.it/}} and TAASP 2022\footnote{\url{http://www.kr.tuwien.ac.at/events/taasp22/}} as well as the ICLP 2023 reviewers for their valuable suggestions,
	to the TU Wien for providing access to computational resources for running the experiments and to Tobias Geibinger for helping with using this infrastructure,
	and to Mark Law for helping with using \slv{FastLAS}.
}

\bibliography{learningdomspecheu}
\bibliographystyle{eptcsini}

\end{document}